%% file: main.tex
\documentclass[sigconf]{acmart}

\usepackage{flushend}
\usepackage{balance}

\def\BibTeX{{\rm B\kern-.05em{\sc i\kern-.025em b}\kern-.08emT\kern-.1667em\lower.7ex\hbox{E}\kern-.125emX}}

\acmSubmissionID{9594}

\usepackage{multirow}
\usepackage{booktabs}
\usepackage{graphicx}
\usepackage{subfigure} 
\usepackage{url}       
\usepackage{stfloats}  
\usepackage{amsmath}   
\usepackage{array}
\usepackage{breqn}
\usepackage{amsfonts}
\usepackage{courier}

\usepackage{amssymb}
\usepackage{balance}
\usepackage{multirow,tabularx}
\usepackage{longtable}
\usepackage{booktabs,longtable}
\usepackage{hhline}
\usepackage{dsfont}
\usepackage{fancyhdr}
\usepackage{epstopdf}
\usepackage{algorithm}
\usepackage{enumitem}
\usepackage{algorithmic}
\usepackage[bottom]{footmisc}
\usepackage{verbatim}
\usepackage{comment}

\newcolumntype{L}[1]{>{\raggedright\let\newline\\\arraybackslash\hspace{0pt}}m{#1}}
\newcolumntype{C}[1]{>{\centering\let\newline\\\arraybackslash\hspace{0pt}}m{#1}}
\newcolumntype{R}[1]{>{\raggedleft\let\newline\\\arraybackslash\hspace{0pt}}m{#1}}

\newfont{\mycrnotice}{ptmr8t at 7pt}
\newfont{\myconfname}{ptmri8t at 7pt}

\copyrightyear{2024}
\acmYear{2024}
\setcopyright{acmlicensed}
\acmConference[SIGIR '24]{Proceedings of the 47th International ACM SIGIR Conference on Research and Development in Information Retrieval}{July 14--18, 2024}{Washington, DC, USA}
\acmBooktitle{Proceedings of the 47th International ACM SIGIR Conference on Research and Development in Information Retrieval (SIGIR '24), July 14--18, 2024, Washington, DC, USA}
\acmISBN{979-8-4007-0431-4/24/07}
\acmDOI{10.1145/3626772.3657831}

\begin{document}

\title{Fine-grained Textual Inversion Network for Zero-Shot Composed Image Retrieval}

\author{Haoqiang Lin}
\orcid{0009-0000-5768-5467}
\affiliation{%
	\institution{Shandong University}
	  \city{Qingdao}\country{China}
}
\email{zichaohq@gmail.com}

\author{Haokun Wen}
\orcid{0000-0003-0633-3722}
\affiliation{%
	\institution{\mbox{Harbin Institute of Technology~(Shenzhen)}}
	  \city{Shenzhen}\country{China}
}\email{whenhaokun@gmail.com}

\author{Xuemeng Song*}
\orcid{0000-0002-5274-4197}
\affiliation{%
	\institution{Shandong University}
	  \city{Qingdao}\country{China}
}\email{sxmustc@gmail.com}

\author{Meng Liu}
\orcid{0000-0002-1582-5764}
\affiliation{%
	\institution{Shandong Jianzhu University}
	  \city{Jinan}\country{China}
}\email{mengliu.sdu@gmail.com}

\author{Yupeng Hu}
\orcid{0000-0002-5653-8286}
\affiliation{%
	\institution{Shandong University}
	  \city{Jinan}\country{China}
}\email{huyupeng@sdu.edu.cn}

\author{Liqiang Nie}
\orcid{0000-0003-1476-0273}
\affiliation{%
	\institution{\mbox{Harbin Institute of Technology~(Shenzhen)}}
	 \city{Shenzhen}\country{China}
}\email{nieliqiang@gmail.com}
\renewcommand{\shortauthors}{Haoqiang Lin et al.}

\thanks{*Xuemeng Song (sxmustc@gmail.com) is the corresponding author.}
\begin{abstract}
Composed Image Retrieval (CIR) allows users to search target images with a multimodal query, comprising a reference image and a modification text that describes the user's modification demand over the reference image. 
Nevertheless, due to the expensive labor cost of training data annotation, recent researchers have shifted to the challenging task of zero-shot CIR (ZS-CIR), which targets fulfilling CIR without annotated triplets. 
The pioneer ZS-CIR studies focus on converting the CIR task into a standard text-to-image retrieval task by pre-training a textual inversion network that can map a given image into a single pseudo-word token. 
Despite their significant progress, their coarse-grained textual inversion may be insufficient to capture the full content of the image accurately. 
To overcome this issue, in this work, we propose a novel Fine-grained Textual Inversion Network for ZS-CIR, named FTI4CIR. In particular, FTI4CIR comprises two main components: fine-grained pseudo-word token mapping and tri-wise caption-based semantic regularization. 
The former maps the image into a subject-oriented pseudo-word token and several attribute-oriented pseudo-word tokens to comprehensively express the image in the textual form, while the latter works on jointly aligning the fine-grained pseudo-word tokens to the real-word token embedding space based on a BLIP-generated image caption template. 
Extensive experiments conducted on three benchmark datasets demonstrate the superiority of our proposed method. 
\end{abstract}



\maketitle

\input{sec_intro}

\input{sec_rel}

\input{sec_meth}

\input{sec_exper}

\input{sec_concl}

\begin{acks}
This work was supported in part by the National Natural Science Foundation of China, No.: 62376137, No.:62276155, No.:62376140 and No.:U23A20315; 
the Shandong Provincial Natural Science Foundation, No.:ZR2022YQ59, and No.:ZR2021MF040; 
the Science and Technology Innovation Program for Distinguished Young Scholars of Shandong Province Higher Education Institutions, No.: 2023KJ128.

\end{acks}


\bibliographystyle{ACM-Reference-Format}  
\balance
\bibliography{sigproc_abbre}


\end{document}

%% file: sec_intro.tex
\label{sec: intro}

\begin{figure}[!t]
		\centering
		\includegraphics[scale=0.53]{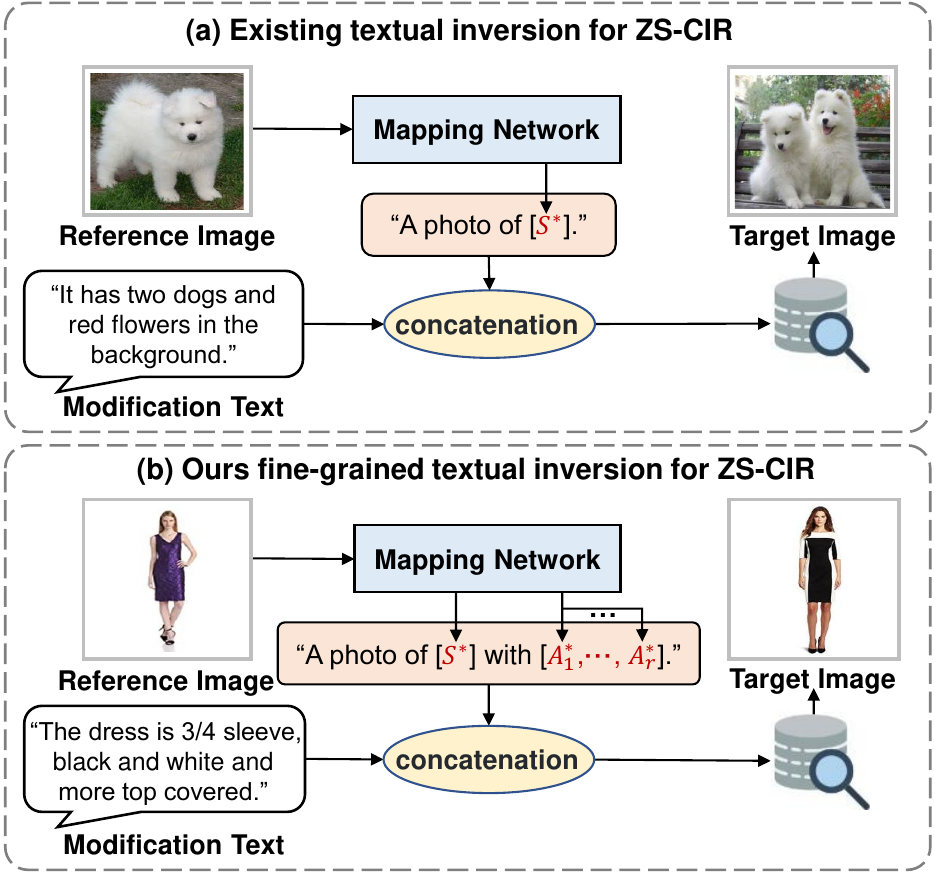}
	    \caption{An illustration of method comparison. (a) Exiting textual inversion for ZS-CIR. (b) Our fine-grained textual inversion for ZS-CIR.}\label{diff_comparasion}
     \vspace{-1.5em}
\end{figure}

\section{Introduction}

Unlike traditional \mbox{content-based} image retrieval~\cite{chua1994content,qu2021dynamic} or \mbox{text-based} image retrieval~\cite{qu2020context,rao2022does}, composed image retrieval (CIR) enables users to search the target image with a multimodal query: a reference image plus a modification text that indicates their desired modification demand over the reference image. In light of its significant application potential in e-commerce and internet search engines~\cite{jandial2022sac,wu2021fashion}, CIR has garnered considerable research attention in recent years~\cite{lee2021cosmo,yuan2021conversational,chen2020learning,wen2023self,hosseinzadeh2020composed}. Current CIR methods mainly fall into the supervised learning paradigm. Namely, they heavily rely on annotated triplets in the form of \textless \emph{reference image, modification text, target image}\textgreater. However, it is time-consuming to annotate the modification text for each potential \textless \emph{reference image, target image}\textgreater\hspace{0.2em}pair. Therefore, existing CIR datasets suffer from limited scale, constraining the generalization ability of current supervised methods, especially for queries from unseen domains.

To eliminate the model's dependency on labeled datasets, recent researches~\cite{saito2023pic2word,Baldrati_2023_ICCV} have introduced the new task of \mbox{zero-shot} CIR (\mbox{ZS-CIR}), which aims to address CIR without requiring any annotated training triplet. As depicted in Figure~\ref{diff_comparasion}~(a), inspired by the textual inversion technique~\cite{wei2023elite,gal2023an,cohen2022my}, current ZS-CIR solutions focus on training a mapping network over a frozen vision-language pre-trained model, such as CLIP~\cite{radford2021learning}, which can convert the image's visual embedding into a latent pseudo-word token.
This pseudo-word token should possess two important properties: a) encapsulating the informative content of the image, and b) being compatible with the textual token embedding space of real-words. In this manner, each image can be represented by the sentence ``a photo of $S^*$'' and gains a textual embedding, where $S^*$ is the to-be-learned pseudo-word. 
Notably, the mapping network is trained simply based on a set of unlabeled open-domain real-world images instead of the labeled triplets.
During inference, when provided with a composed query, the reference image can be converted into a sentence with a pseudo-word and hence be seamlessly integrated into the token sequence of the given modification text, forming a unified text query. Ultimately, the CIR task can be fulfilled with a standard text-to-image retrieval model. 

Although significant progress has been made in ZS-CIR studies, they primarily concentrate on converting the input image into a general pseudo-word token based on the image's global visual feature. However, these approaches may not accurately capture the full content of the image. 
In CIR tasks, users' modification requests generally fall into subject-oriented and attribute-oriented. Subject-oriented requests involve altering the category or number of the primary subject(s) in the image, while attribute-oriented requests pertain to changing the attributes (e.g., \textit{background} and \textit{sleeve length}) of these primary subject(s). Therefore, we propose conducting fine-grained textual inversion to better address the downstream CIR tasks. As shown in Figure ~\ref{diff_comparasion}~(b), we map each image into a subject-oriented pseudo-word token and several attribute-oriented pseudo-word tokens to express the image fully. 
Then each image would be represented by the sentence ``a photo of [$S^*$] with [[${A_1^*,\cdots, A_r^*}$]]'', where $S^*$ refers to the subject-oriented pseudo-word encapsulating the primary subject(s) information of the image, while $A_i^*$ $(i=1,\cdots,r)$ stand for the attribute-oriented pseudo-words containing the contextual attributes of the primary subject(s).

However, this is \mbox{non-trivial} owing to the following two challenges. 
\textbf{1)} \textbf{Images of different domains usually involve diverse local attributes.} For instance, images from the fashion domain typically contain the attributes of \textit{sleeve length}, \textit{waist design}, and \textit{color}, while images from the animal domain usually contain the attributes of \textit{background}, \textit{position}, and \textit{fur}. Therefore, how to effectively capture the diverse local attributes across different images poses a crucial challenge. 
\textbf{2)} \textbf{Jointly regulate the projection of both subject-oriented and attribute-oriented pseudo-word tokens.} Previous ZS-CIR work has adopted the image-related categories as real-word tokens to promote the projected pseudo-word token to reside in the real-word token embedding space. However, we argue that simply using the general image categories is insufficient to fully regularize the projection of both subject-oriented and attribute-oriented pseudo-word tokens. Consequently, how to jointly align subject-oriented and attribute-oriented pseudo-word tokens to the real-word token embedding space forms another challenge.

To address these challenges, we propose a Fine-grained Textual Inversion Network for Zero-Shot Composed Image Retrieval, dubbed FTI4CIR. As shown in Figure~\ref{fig:model_struct}, our FTI4CIR consists of two key components: fine-grained pseudo-word token mapping and tri-wise caption-based semantic regularization. 
The first component works on mapping the image into a subject-oriented pseudo-word token and several attribute-oriented pseudo-word tokens based on the image's global feature and local attribute features, respectively. 
In particular, to cope with the diverse types of local attributes across images in different domains, we devise the dynamic local attribute feature extraction module, which first uncovers all the possible local attribute features by adaptively aggregating the image's local patch features by Transformer~\cite{vaswani2017attention} and then filters out the learned irrelevant attribute features with a local-global relevance-based filtering strategy. 
The second component targets jointly aligning the subject-oriented and attribute-oriented pseudo-word tokens to the real-word token embedding space. Towards this end, we first employ BLIP~\cite{li2022blip} to generate the real-word description for each image, which is typically in the format of ``[primary subject(s)] + [detailed description]''. We then design three text templates based on the generated caption to facilitate a tri-wise (namely, subject-wise, attribute-wise, and whole-wise) caption-based semantic regularization. 
By pushing the embedding of the pseudo-word-involved template to be close to that of the original real-word caption, the tri-wise caption-based semantic regularization can promote not only the interaction between the pseudo-words and their counterpart real-words, but also that between the pseudo-words and the other contextual real-words in the image caption.

In the inference phase, given a multimodal query of the CIR task, FTI4CIR first converts the reference image into a sentence with pseudo-words, and then concatenates it with the modification text to derive a final pure text query, thereby simplifying the CIR task to a standard text-to-image retrieval task. 
Our main contributions can be summarized as follows:
\begin{itemize}
    \item To the best of our knowledge, we are the first to explore the fine-grained textual inversion, \textit{i.e.}, mapping the image into a subject-oriented pseudo-word token and several attribute-oriented pseudo-word tokens to realize ZS-CIR. 
    \item To facilitate the attribute-oriented pseudo-word token mapping, we propose a dynamic local attribute feature extraction module to handle the problem of diverse types of local attributes across images in different domains.
 \item We devise a tri-wise caption-based semantic regularization that enables the thorough interactions between pseudo-words and real-words, and hence promotes aligning the pseudo-word tokens to the real-word token embedding space. We have released our codes to facilitate other researchers\footnote{\url{https://github.com/ZiChao111/FTI4CIR}.}.

\end{itemize}

%% file: sec_rel.tex
\begin{figure*}[!t]
    \centering
    \includegraphics[scale=0.56]{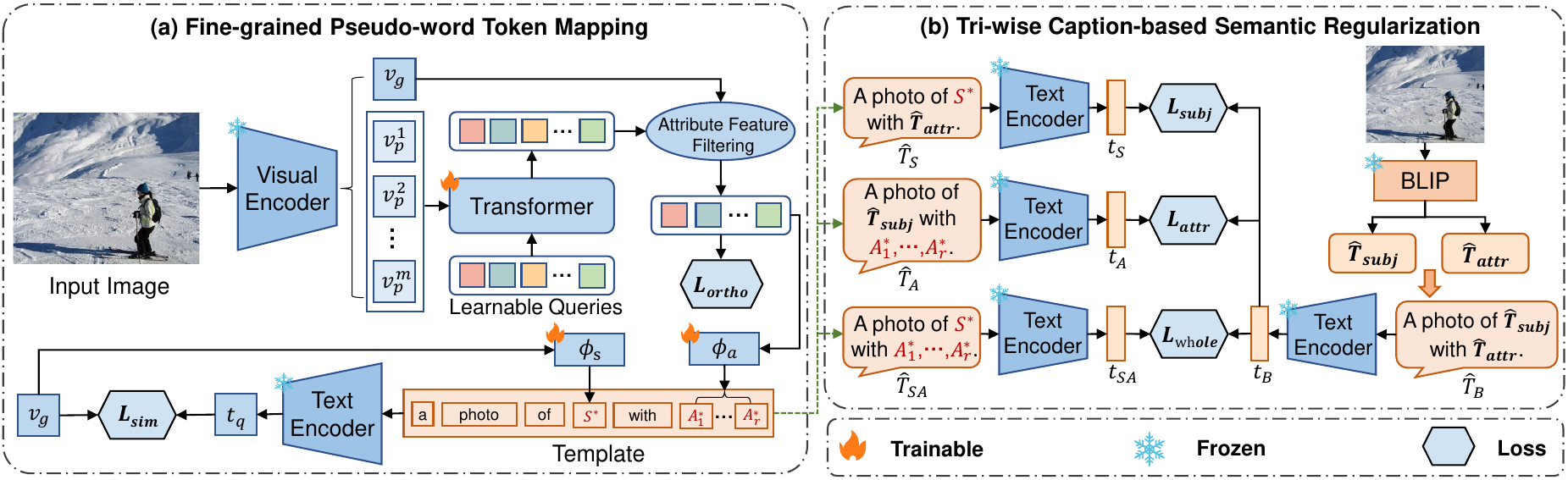}
    
    \caption{The proposed FTI4CIR consists of two key modules: (a) Fine-grained pseudo-word token mapping and (b) Tri-wise caption-based semantic regularization.}\label{fig:model_struct}
\end{figure*}


\section{Related Work}
\label{sec: rel}
Our work is closely related to composed image retrieval and vision-language pre-trained models. 

\textbf{Composed Image Retrieval.} 
Based on the type of feature extraction backbone, existing CIR methods can be primarily divided into two groups. The former group~\cite{vo2019composing,lee2021cosmo,wen2021comprehensive,gu2021image,kim2021dual,anwaar2021compositional} mainly utilizes the traditional models, such as ResNet~\cite{he2016deep} and LSTM~\cite{hochreiter1997long}, to extract image and text features, respectively. In contrast, the second group~~\cite{baldrati2022effective,baldrati2022conditioned,wen2023target,goenka2022fashionvlp,lin2023clip,han2023fame} takes advantage of vision-language pre-trained (VLP) models, like CLIP, for feature extraction. 
Typically, these VLP model-based methods usually yield better performance due to their superior feature extraction ability on multimodal data. Nevertheless, all the above approaches fall into the supervised learning paradigm, requiring costly annotated triplets to tune the model. Hence, recent researchers have shifted the ZS-CIR task to fulfill CIR without labeled triplets. For example, Saito \textit{et al.}~\cite{saito2023pic2word} employed a pre-trained textual inversion network to convert the reference image into a single pseudo-word token within the CLIP token embedding space. This pseudo-word token is then seamlessly integrated into the token sequence of the given modification text to retrieve the target image.
Meanwhile, Baldrati \textit{et al.}~\cite{Baldrati_2023_ICCV} respectively designed an optimization-based textual inversion method and a mapping network-based method to learn a pseudo-word token for encapsulating the visual content of each image. Both methods involve category-based semantic regularization to align the pseudo-word token to the CLIP token embedding space. Although these ZS-CIR studies have achieved significant progress, they focus on using a single pseudo-word to represent the whole information of each image, which may miss the rich detailed information contained in the image. In light of this, we propose conducting the fine-grained textual inversion, \textit{i.e.}, mapping the image into a subject-oriented pseudo-word token and several attribute-oriented pseudo-word tokens, to fully express the image in the textual form.

\textbf{Vision-language Pre-trained Models.} 
Recently, VLP models have attracted considerable research attention. The standard VLP models, like CLIP and ALIGN~\cite{li2021align}, are pre-trained on large-scale image-text pairs to acquire knowledge of implicit alignment between images and texts. Due to the powerful cross-modal alignment and feature extraction capability, they have seen wide-ranging usage from diverse tasks like zero-shot classification~\cite{radford2021learning}, fine-grained classification~\cite{conde2021clip}, and video-text retrieval~\cite{liu2022animating}. Further, beyond simple as a feature extraction backbone, there have been many attempts to unify various vision and language tasks into a single framework of the VLP model~\cite{wang2022simvlm,cho2021unifying}. For example, BLIP is a multimodal mixture of encoder-decoder VLP model that has demonstrated remarkable performance across understanding-based tasks (e.g., visual question answering) and generation-based tasks (e.g., image captioning). 
In this study, we employed the standard VLP model CLIP as the feature extraction backbone for our model to realize fine-grained textual inversion. Additionally, we utilized BLIP as an image captioning model to generate the image's caption to align the projected pseudo-word tokens into the real-word token embedding space.

%% file: sec_meth.tex
\vspace{-1.0em}
\section{FTI4CIR}
\label{sec: meth}
To address the task of ZS-CIR, we propose a Fine-grained Textual Inversion Network for Zero-Shot Composed Image Retrieval, named FTI4CIR, which can be trained simply on an unlabeled image set. As shown in Figure~\ref{fig:model_struct}, FTI4CIR consists of two key components: fine-grained pseudo-word token mapping and tri-wise caption-based semantic regularization. 
As a major novelty, different from existing work that maps an image into a coarse global pseudo-word token, FTI4CIR maps the image into not only a subject-oriented pseudo-word token $\mathbf{s}$ but also several attribute-oriented pseudo-word tokens $[\mathbf{a_1},\cdots, \mathbf{a_r}]$. 
For optimization, we first adopt the commonly used contrastive loss to correlate the original visual embedding and the pseudo-word-based textual embedding of the given image. In addition, we devise a tri-wise caption-based semantic regularization, which fully utilizes BLIP-generated high-quality image captions to model the thorough interactions between pseudo-words and real-words, and hence promotes the pseudo-word tokens learning. Once our FTI4CIR gets well-trained, it can be applied to various downstream CIR tasks in the inference phase, where the multimodal CIR query would be unified into a pure text query and hence simplify each CIR task into a standard text-to-image retrieval task.

\subsection{Fine-grained Pseudo-word Token Mapping}
In order to encapsulate the image's content in a fine-grained manner, we project the image into both subject-oriented and attribute-oriented pseudo-word tokens. The former captures the primary subject(s) and the latter captures its/their associated attributes.

\textbf{Subject-oriented Pseudo-word Token Mapping}. Intuitively, the subject-oriented pseudo-word token should capture the primary subject(s) information of the image. To achieve this, we leverage the widely adopted CLIP visual encoder, inspired by its remarkable success in CIR tasks~\cite{baldrati2022effective, wen2023target}, to extract the global feature of the input image. Similar to previous works, we regard the last-layer output of the frozen CLIP visual encoder as the global feature of the image, denoted as ${ \mathbf{v}_g \in \mathbb{R}^{d_1}}$, where $d_1$ represents the dimension of the global feature. 
We then employ a Multi-layer perceptron (MLP), acting as a mapping network, to project the global image feature into a subject-oriented pseudo-word token. Formally, let $\phi_{s}$ be the mapping function. We then have:
\begin{equation}
\mathbf{s} = {\phi_{s}} \left({\mathbf{v}_g}\right), 
\end{equation}
where $\mathbf{s}$ denotes the subject-oriented pseudo-word token.

\textbf{Attribute-oriented Pseudo-word Token Mapping}. 
The attribute-oriented pseudo-word tokens are intended to encapsulate the local detailed attribute information of the primary subject(s) in the image. 
Therefore, for deriving the attribute-oriented pseudo-word tokens, we resort to the image's local features. Recent advancements in VLP models have made the use of local patch features quite prevalent. 
Nevertheless, individual patch features do not directly represent the local semantic attributes, as each attribute is usually correlated with multiple patches~\cite{wen2023target}. For example, the \textit{collar design} attribute is typically associated with the top patches of an image, while the \textit{sleeve length} attribute is connected to patches on both the left and right sides.

\textit{Unified Local Attribute Features Extraction.} Accordingly, to facilitate the derivation of attribute-oriented pseudo-word tokens, we propose extracting the local attribute features of each image by adaptively aggregating the patch features. As mentioned above, the types of attributes usually vary across images from different domains. Therefore, we propose the dynamic local attribute feature extraction module. In particular, we first assume that there are a total of $n$ latent local attributes present in all open-domain real-world images. We then learn these local attribute features for each image by adaptively aggregating its local patch features. To mitigate the adverse impact of the learned noisy irrelevant attributes (e.g., the \textit{collar length} attribute for an image with natural scenery), we introduce a local-global relevance-based filtering strategy to retain only the relevant local attribute features for subsequent attribute-oriented pseudo-word tokens learning. 

Formally, let $\mathbf{V}=\left\{\mathbf{v}_p^i\right\}_{i=1}^m \in \mathbb{R}^{d_2 \times m}$ represent the image patch features, output by the second-to-last layer of the frozen CLIP visual encoder. $d_2$ denotes the dimension of the patch features and $m$ denotes the number of image patches. Then for local attribute features extraction, we adopt the \mbox{off-the-shelf} Transformer Network, which can model a comprehensive interaction among the inputs by the self-attention mechanism. Specifically, we define a set of $n$ learnable query embeddings $\mathbf{X}=\left\{\mathbf{x}_i\right\}_{i=1}^n \in \mathbb{R}^{d_2 \times n}$, where each learnable query embedding corresponds to a specific latent attribute of images. We then feed the concatenation of the learnable query embeddings $\mathbf{X}$ and all the image patch features $\mathbf{V}$ into the Transformer encoder. We regard the outputted corresponding query embeddings as the local attribute features of the image. To facilitate the subsequent local-global relevance-based filtering, we also introduce a fully connected network to ensure that the dimension of the local attribute feature matches that of the global image feature. Formally, this process can be expressed as follows,
\begin{equation}
\mathbf{X}^{\prime}=FC\left(\mathcal{F}_{\text {Transformer }}\left( \left[  \mathbf{X} | \mathbf{V}  \right] \right)\right),
\label{eq:Transformer}
\end{equation}
where $\left[ \cdot | \cdot \right]$ represents cascade operation, $FC$ denotes the fully connected network, and $\mathbf{X}^{\prime} =\left\{\mathbf{x}_i^{\prime}\right\}_{i=1}^n \in \mathbb{R}^{d_1 \times n}$ denotes the local attribute features of the image.

\textit{Local-global Relevance-based Filtering}. As mentioned above, we need to filter out the irrelevant local attribute features of the given image to enhance the interaction between the reference image and modification text, in the downstream CIR tasks. For this purpose, we propose the local-global relevance-based filtering strategy, where the reliable global image feature is used as the reference for selecting the relevant local attribute features. Essentially, the more similar a given local attribute feature is to the global image feature, the more likely it is to be relevant to the given image. Specifically, we identify the relevant local attribute feature by jointly considering its relative similarity ranking and absolute similarity score. Firstly, we rank the learned local attributes features according to their cosine similarities to the given global image feature as follows,
\begin{equation}
\left\{
\begin{array}{ll}
c_i = cos(\mathbf{x}_{i}^{\prime}, \mathbf{v}_g), i=1, \cdots, n,  \\
\mathbf{W} = \left[\mathbf{x}_{j}^{\prime}\right], c_j \in {\text{top-}k\left(\mathcal{C}\right)}, 
\end{array}
\right.
\label{eq:select_topk_features}
\end{equation}
where $cos(\cdot, \cdot)$ denotes the cosine similarity, $c_i$ indicates the similarity between the global image feature $\mathbf{v}_g$ and the $i$-th learned local attribute feature $\mathbf{x}_{i}^{\prime}$. $\text{top-}k(\mathcal{C})$ denotes the top $k$ entry values of $\mathcal{C}$, where $\mathcal{C} = \left[c_1, c_2, \cdots, c_n\right]$. $\mathbf{W} \in \mathbb{R}^{d_1 \times k}$ represents the selected valid local attribute features.

We then set a similarity threshold to further guarantee the quality of the retained local attribute features as follows,
\begin{equation}
\mathbf{W}^{\prime} = \left[\mathbf{x}_{j}^{\prime}\right], c_j \geq \varepsilon \text{ and } \mathbf{x}_{j}^{\prime} \in \mathbf{W},
\label{eq:select_valid_features}
\end{equation}
where $\varepsilon$ is the local-global similarity threshold for retaining the final relevant local attribute features. $\mathbf{W}^{\prime} \in \mathbb{R}^{d_1 \times r}$ represents the set of final selected local attribute features\footnote{When there are no local attribute features that satisfy our filtering criteria, we preserve the most correlated one in $\mathcal{C}$ to maintain the model integrity.}, $r \in \left[1,k\right]$ is the number of selected local attribute features.

Notably, to guarantee that different local attribute features can indeed represent different visual attributes, we introduce an orthogonal loss that enforces the distinctiveness of each feature. In particular, we deploy the orthogonal loss over $\mathbf{W}$ rather than $\mathbf{W}^{\prime}$ to ensure the discrimination between the image's local attribute features as much as possible, while preventing interference from irrelevant features (low similarity features), thus facilitating the local attribute features learning. It can be formulated as follows,
\begin{equation}
{\mathcal{L_{\text{ortho}}}} = \left\|\mathbf{W}  \mathbf{W}^{\top} - \mathbf{I}\right\|_{F}^{2}, 
\label{eq:orth_loss}
\end{equation}
where $\mathbf{I}$ represent the identity matrix, and $\left\| \cdot \right\|_{F}$ refers to the Frobenius norm of the matrix. 

\textit{Mapping.} Given the inherent differences between the global image feature and local attribute features, we employ another MLP-based mapping network, denoted as $\phi_{l}$, to project the local attribute features into the real-word token embedding space. Mathematically, we have: 
\begin{equation}
[\mathbf{a}_1,\cdots,\mathbf{a}_r] = {\phi_{a}\left(\mathbf{W}^{\prime}\right)},
\end{equation}
where $\mathbf{a}_i$ is the $i$-th mapped attribute-oriented pseudo-word token. 

\textbf{Contrastive Cross-modal Learning.}
Having acquired the subject-oriented and attribute-oriented pseudo-word tokens, we can represent each image in the textual form. Specifically, we design the pseudo-word-based textual template: ``a photo of [$S^*$] with [${[A_1^*,\cdots, A_r^*]}$].'' to represent each image, where $S^*$ is the pseudo-word corresponding to the subject-oriented pseudo-word token $\mathbf{s}$, and $A_i^*$ is that corresponding to the $i$-th attribute-oriented pseudo-word token $\mathbf{a}_i$. Accordingly, by feeding this template into the frozen CLIP text encoder, we can obtain the pseudo-word-based textual representation, denoted as $\mathbf{t}_q$, for each image. 

To supervise the pseudo-word tokens learning with the unlabeled pre-training image dataset, following previous work, we adopt the symmetric contrastive loss~\cite{Baldrati_2023_ICCV}. Intuitively, we expect that for each unlabeled image, its pseudo-word-based textual representation should be more aligned with its original visual representation than the visual representations of other images, and vice versa. This loss can be written as follows,
\begin{equation}
\begin{aligned}
\mathcal{L}_{\text {sim}} = & -\frac{1}{B} \sum_{i=1}^B 
\Bigg\{
\log \frac{e^{(\cos(\mathbf{v}_g^i, \mathbf{t}_q^i)/\tau)}}{\sum_{j=1}^B e^{(\cos(\mathbf{v}_g^i, \mathbf{t}_q^j)/\tau)} + \sum_{j \neq i} e^{(\cos(\mathbf{t}_q^i, \mathbf{t}_q^j)/\tau)}} \\
&+ \log \frac{e^{(\cos(\mathbf{t}_q^i, \mathbf{v}_g^i)/\tau)}}{\sum_{j=1}^B e^{(\cos(\mathbf{t}_q^i, \mathbf{v}_g^j)/\tau)} + \sum_{j \neq i} e^{(\cos(\mathbf{v}_g^i, \mathbf{v}_g^j)/\tau)}}
\Bigg\},
\end{aligned}
\label{eq:simCLR_loss}
\end{equation}
where ${B}$ is the batch size, and $\tau$ is a temperature hyperparameter. $\mathbf{v}_g^i$ and $\mathbf{t}_q^i$ represent the global image feature and the pseudo-word-based textual representation of the $i$-th image, respectively.

\subsection{Tri-wise Caption-based Semantic Regularization}
To align the pseudo-word tokens to the real-word token embedding space for facilitating the subsequent combination of the reference image and the modification text embeddings in the inference phase, we follow previous work~\cite{Baldrati_2023_ICCV} and further regularize the interaction between pseudo-word tokens and real-word tokens. The existing solution~\cite{Baldrati_2023_ICCV} focuses on aligning the pseudo-word token with the image-related categories. In our context, it is apparent that simply using image-related categories to guide the subject-oriented and attribute-oriented pseudo-word tokens learning can lead to sub-optimal performance. This is because the image-related category tokens only convey information about the image's category, lacking detailed specifics such as the number of primary subject(s) and local attributes present in the image. 

\begin{figure}[!t]
		\centering
		\includegraphics[scale=0.46]{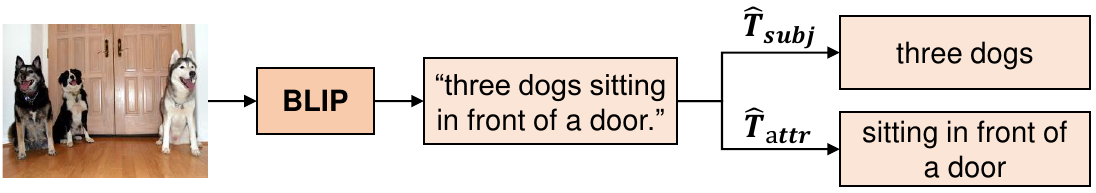}
  \vspace{-1.5em}
	    \caption{An example of BLIP generated caption, which can be divided into two parts: $\hat{T}_{subj}$ and $\hat{T}_{attr}$.}\label{fig:diff_Sentence}
     \vspace{-0.5em}
\end{figure}

Inspired by the remarkable success of the VLP models in image captioning, we use BLIP to generate a high-quality description for each image to guide the pseudo-word tokens learning. This is because BLIP, pre-trained on the COCO~\cite{lin2014microsoft} dataset, typically generates the image caption in the format of ``[primary subject(s)] $+$ [detailed description]'', such as ``[three dogs] [sitting in front of a door].''. As can be seen, this format matches our designed text template for representing each image well. 
To facilitate the following tri-wise caption-based semantic regularization, we first split the generated caption into two parts: one describes the primary subject(s) and the other one delivers the local attributes. Specifically, given the generated caption $\hat{T}$, we employ the POS tagger of the spacy library\footnote{\url{https://spacy.io}.} to identify its first subject term. Then we denote the former sub-sequence of $\hat{T}$ ending with the detected subject term as $\hat{T}_{subj}$, and the remaining sub-sequence as $\hat{T}_{attr}$.
As illustrated in Figure~\ref{fig:diff_Sentence}, given the example description ``three dogs sitting in front of a door.'', we can split it into two parts: $\hat{T}_{subj}$ refers to ``three dogs'' describing the primary subjects of the image, while $\hat{T}_{attr}$ refers to ``sitting in front of a door'' conveying the local attribute information about the primary subjects. 

One naive approach to regulating the pseudo-word tokens learning is to model the interaction between the pseudo-words and the counterpart real-words. Namely, pushing the subject-oriented pseudo-word $S^*$ to be close to $\hat{T}_{subj}$, while the attribute-oriented pseudo-words $A_i^*$ $(i=1,\cdots,r)$ to $\hat{T}_{attr}$. However, this approach only considers the local interaction, but ignores the interaction between the pseudo-words and other contextual real-words, such as the interaction between $S^*$ and $\hat{T}_{attr}$, and that between $A_i^*$ $(i=1,\cdots,r)$ and $\hat{T}_{subj}$. In fact, due to the sequence nature of the text, modeling the interaction between the pseudo-words and other contextual real-words should be also beneficial for the CLIP text encoder to interpret the pseudo-word-based textual form of each image. Accordingly, we propose to conduct the tri-wise caption-based semantic regularization within the context. 

Notably, to unify the caption format and the aforementioned pseudo-word-based textual template, we standardize each generated caption with the text template ``a photo of [$\hat{T}_{subj}$] with [$\hat{T}_{attr}$].''. Let $\hat{T}_{B}$ denotes the standard caption of the image. Based on this caption, we then derive three derivatives: $\hat{T}_S$, $\hat{T}_A$, and $\hat{T}_{SA}$ in Figure~\ref{fig:model_struct}~(b). In particular, $\hat{T}_S$ replaces $\hat{T}_{subj}$ with the subject-oriented pseudo-word $S^*$, $\hat{T}_A$ replaces $\hat{T}_{attr}$ with the attribute-oriented pseudo-words $A_i^*$ $(i=1,\cdots,r)$, and $\hat{T}_{SA}$ replaces both. Based on the standard caption and its three derivatives, we can conduct a tri-wise caption-based semantic regularization as follows,
\begin{equation}
\left\{
\begin{array}{ll}
{\mathcal{L_{\text{subj}}}} &\!\!\!\!\!= 1- \cos({t_{B}}, {t_{S}}), \\
{\mathcal{L_{\text{attr}}}} &\!\!\!\!\!= 1- \cos({t_{B}}, {t_{A}}), \\
{\mathcal{L_{\text{whole}}}} &\!\!\!\!\!= 1- \cos({t_{B}}, {t_{SA}}), \\
{\mathcal{L_{\text{tri-reg}}}} &\!\hspace{0.018em}\!\!\!\!= {\mathcal{L_{\text{subj}}}} + {\mathcal{L_{\text{attr}}}} + {\mathcal{L_{\text{whole}}}}, 
\end{array}
\right.
\end{equation}
where ${t_{B}}$, ${t_{S}}$, ${t_{A}}$, and ${t_{SA}}$ are the embedding derived by the frozen CLIP text encoder for $\hat{T}_{B}$, $\hat{T}_S$, $\hat{T}_A$, and $\hat{T}_{SA}$, respectively. 
Losses ${\mathcal{L_{\text{subj}}}}$ and ${\mathcal{L_{\text{attr}}}}$ are specifically designed to guide the learning of the subject-oriented pseudo-word token and attribute-oriented pseudo-word tokens, respectively. 
${\mathcal{L_{\text{whole}}}}$ is devised to jointly regulate all the pseudo-word tokens learning. Essentially, our goal is to ensure that the pseudo-word-based caption derivatives are semantically close to the real-word-based caption.

The final loss used to optimize FTI4CIR is as follows,
\begin{equation}
\mathcal{L_{\text{total}}} = \mathcal{L_{\text{sim}}} + \mathcal{L_{\text{ortho}}} + {\lambda}_{\text{reg}}  \mathcal{L_{\text{tri-reg}}} , 
\label{eq:total_loss}
\end{equation}
where ${\lambda}_{\text{reg}}$ is the trade-off hyper-parameters.


\subsection{Inference with Pre-trained FTI4CIR}
During the inference phase, we combine the reference image $I_r$ and its modification text $T_m$ as a composed query to retrieve the target image $I_t$. 
To be specific, we first employ the well-trained FTI4CIR to comprehensively map the reference image into both subject-oriented and attribute-oriented pseudo-word tokens to represent the image in the textual form. Then similar to previous work, we introduce the template ``a photo of [$S^*$] with [${[A_1^*,\cdots, A_r^*]}$] but [$T_m$].'', to compose the reference image and modification text into a sentence. Thereafter, we encode the composed query by CLIP text encoder and each candidate image by CLIP visual encoder, and based on the corresponding output embeddings to measure their similarity. Ultimately, we rank the candidate images according to their similarities.

%% file: sec_exper.tex
\section{Experiment}
\label{sec:exper}

In this section, we first introduce the experimental settings and then provide the experiment results as well as corresponding analyses to answer the following research questions:
\begin{itemize}
    \item \textbf{RQ1.} Does FTI4CIR surpass existing methods?
    \item\textbf{RQ2.} How does each component affect FTI4CIR?
    \item \textbf{RQ3.} Is \mbox{FTI4CIR} sensitive to the key hyperparameter?
    \item\textbf{RQ4.} How is the qualitative performance of FTI4CIR?
\end{itemize}

\subsection{Experimental Setting}

\subsubsection{Evaluation Dataset.} To evaluate the performance of our FTI4CIR in various downstream CIR tasks, following~\cite{Baldrati_2023_ICCV}, we chose three public datasets for evaluation: FashionIQ~\cite{wu2021fashion}, CIRR~\cite{liu2021image}, and CIRCO~\cite{Baldrati_2023_ICCV}. 
\textbf{FashionIQ} contains fashion items of three categories: Dresses, Shirts, and Tops\&Tees. Similar to previous studies~\cite{zhao2022progressive}, since the test set remains undisclosed, we evaluated our model on the validation set, which in total consists of $6$K validation triplets of three categories. \textbf{CIRR} comprises \mbox{$\sim$\hspace{0em}$21$K} real-life open-domain images taken from the NLVR$^{2}$ dataset~\cite{suhr2018corpus}. To avoid false negatives, the annotation process requires that the modifying text is only relevant to one image pair and irrelevant to any other image pairs sharing the same reference image. We assessed our model on the test set of CIRR, which contains $4.1$K testing triplets. \textbf{CIRCO} is an open-domain dataset recently developed from the COCO dataset for further addressing the false negative issue. Different from the above two datasets, in CIRCO, each sample comprises a reference image, a modification text, and multiple target images. We utilized the test set that consists of $800$ samples for evaluating our model.

\begin{table*}
  \centering
  \caption{\textbf{Performance comparison on FashionIQ.} The best results are in boldface, while the second best results are underlined. 
  }
  \label{tab: baseline_FashionIQ}
    \begin{tabular}{c|l|cc|cc|cc|cc}
    \hline
    \multirow{2}{*}{Supervision} & \multirow{2}{*}{Method} & \multicolumn{2}{c|}{Dresses} & \multicolumn{2}{c|}{Shirts} & \multicolumn{2}{c|}{Tops\&Tees} & \multicolumn{2}{c}{Avg} \\
    \cline{3-4} \cline{5-6} \cline{7-8} \cline{9-10}
         & & \textbf{R@$10$}& \textbf{R@$50$}& \textbf{R@$10$}& \textbf{R@$50$}& \textbf{R@$10$}& \textbf{R@$50$}& \textbf{R@$10$}&\textbf{R@$50$} \\
    \hline \hline
    \multirow{8}{*}{ZERO-SHOT} 
    & {Image-only} & $5.35$& $13.93$& $9.91$& $20.80$& $8.31$& $17.70$& $7.86$& $17.48$\\ 
    & Text-only& $14.38$& $32.92$& $19.28$& $33.02$& $21.52$& $39.16$& $18.39$& $35.03$\\
    & Image + Text& $16.81$& $36.14$& $21.10$& $34.49$& $23.97$& $39.42$& $20.62$& $36.69$\\
    & Captioning& $7.98$& $21.76$& $21.49$& $36.16$& $18.77$& $34.17$& $16.08$&  $30.70$\\
    & Pic2Word~\cite{saito2023pic2word}& $20.00$& $40.20$& $26.20$& $43.60$& $27.90$& $47.40$& $24.70$& $43.70$ \\
    &  SEARLE-XL-OTI~\cite{Baldrati_2023_ICCV}& $\underline{21.57}$& $\underline{44.47}$& $\underline{30.37}$& $\underline{47.49}$& $\underline{30.90}$& $\underline{51.76}$& $\underline{27.61}$& $\underline{47.90}$ \\
     &  SEARLE-XL~\cite{Baldrati_2023_ICCV}& $20.48$& $43.13$& $26.89$& $45.58$& $29.32$& $49.97$& $25.56$& $46.23$\\
    & \textbf{FTI4CIR} &   $\mathbf{24.39}$& $\mathbf{47.84}$ & $\mathbf{31.35}$ & $\mathbf{50.59}$ & $\mathbf{32.43}$ & $\mathbf{54.21}$ & $\mathbf{29.39}$ & $\mathbf{50.88}$ \\
    \hline \hline
    FashionIQ & Combiner~\cite{baldrati2022effective}& $30.49$& $54.93$& $37.98$& $57.16$& $38.50$& $60.02$& $35.66$&$57.37$\\
    CIRR & Combiner~\cite{baldrati2022effective}& $20.53$& $40.36$& $25.07$& $43.18$& $26.82$& $47.68$& $24.14$& $43.74$ \\
    \hline
    \end{tabular}%
\end{table*}

\subsubsection{Implementation Details.} 
In order to make a fair comparison, we followed~\cite{Baldrati_2023_ICCV}, adopting the unlabeled test split of ImageNet$1$K~\cite{russakovsky2015imagenet} as the pre-training dataset. This dataset contains $100K$ unlabeled open-domain real-world images with a high variety of subjects. 
We adopted pre-trained CLIP (ViT-L/$14$ version) as the feature extraction backbone of FTI4CIR. The Transformer for deriving the local attribute features of images is set to $3$ layers and $1$ head. We set the number of latent local attributes in local attribute feature learning $n$ to $24$. In terms of the local-global relevance-based filtering, we set the count of valid local attribute features $k$ in~Eqn. ($\ref{eq:select_topk_features}$) and the local-global similarity threshold $\varepsilon$ in~Eqn. ($\ref{eq:select_valid_features}$) to $12$ and $0.05$, respectively. The temperature $\tau$ in Eqn. ($\ref{eq:simCLR_loss}$) is set to $0.2$. Regarding the loss weight, we set ${\lambda}_{\text{reg}}$ in Eqn. ($\ref{eq:total_loss}$) to $1.4$.
We trained FTI4CIR by AdamW~\cite{LoshchilovH19} optimizer with an initial learning rate of $4e-5$. The learning rate decays by a factor of $0.1$ at the $10$-th epoch. 
We empirically set the batch size as $256$. All the experiments are implemented by PyTorch~\cite{paszke2019pytorch}, and we fixed the random seeds to ensure reproducibility. Furthermore, we evaluated the CIR performances of our model in a zero-shot manner, \textit{i.e.}, once FTI4CIR is well-trained with the pre-training image dataset, it can be tested on all three datasets.

\subsubsection{Evaluation.} We used the standard evaluation protocols for each dataset. 
For FashionIQ, consistent with previous studies~\cite{Baldrati_2023_ICCV,saito2023pic2word}, we used the recall at rank $K$ (R@$K$) as the evaluation metric. Specifically, we adopted R@$10$ and R@$50$. In addition, we calculated the average performance across the three subsets of different categories to gauge the overall performance. 
For CIRR, as suggested by previous studies~\cite{liu2021image,delmas2022artemis}, we employed a combination of criteria, namely R@$K$ (where $K$ = $1$, $5$, $10$, $50$), R$_{subset}$@$K$ (where $K$ = $1$, $2$, $3$), and the average of R@$5$ and R$_{subset}$@$1$, as evaluation metrics. 
Notably, R$_{subset}$@$K$ limits the candidate target images to those that are semantically similar to the correct target image to alleviate the problem of false negatives. 
For CIRCO, due to the multiple ground truths, following the previous work~\cite{Baldrati_2023_ICCV}, we adopted Average Precision (mAP) as a more fine-grained metric, specifically mAP@$K$ (where $K$ = $5$, $10$, $25$, $50$).

\begin{table*}
  \centering
  \caption{\textbf{Performance comparison on CIRR.} The best and second-best results are highlighted in bold and underlined, respectively. – denotes results not reported in the original paper. }
  \label{tab: baseline_cirr}
    \begin{tabular}{c|l|cccc|ccc|c}
    \hline
    \multirow{2}{*}{Supervision} & \multirow{2}{*}{Method} & \multicolumn{4}{c|}{\textbf{R@$K$}} & \multicolumn{3}{c|}{\textbf{R$_{subset}$@$K$}} & \multirow{2}{*}{Avg}\\
    \cline{3-9} 
         & & $K$ = $1$ & $K$ = $5$& $K$ = $10$& $K$ = $50$& $K$ = $1$& $K$ = $2$& $K$ = $3$ \\
    \hline \hline
    \multirow{8}{*}{ZERO-SHOT} 
    & Image-only & $7.35$& $23.71$& $33.81$& $57.16$& $20.80$& $42.07$& $61.52$&$22.25$\\ 
    & Text-only& $21.81$& $45.13$& $57.54$& $79.52$& $\mathbf{61.52}$& $\mathbf{80.41}$& $\mathbf{90.31}$&$\underline{53.32}$\\
    & Image + Text& $12.34$& $36.22$& $50.27$& $78.15$& $34.19$& $59.06$& $76.72$&$35.21$\\
    & Captioning& $16.60$& $40.00$& $52.94$& $79.33$& $52.99$& $74.27$& $86.87$&$46.49$\\
    & Pic2Word~\cite{saito2023pic2word}& $23.90$& $51.70$& $65.30$& $87.80$& ---& ---& ---& ---\\
    & SEARLE-XL-OTI~\cite{Baldrati_2023_ICCV}& $\underline{24.87}$& $52.31$& $\underline{66.29}$& $88.58$& ${53.80}$& $74.31$& $86.94$&$53.06$\\
    & SEARLE-XL~\cite{Baldrati_2023_ICCV}& $24.24$& $\underline{52.48}$& $\underline{66.29}$& $\underline{88.84}$& $53.76$& $75.01$& $\underline{88.19}$&$53.12$\\
    & \textbf{FTI4CIR} &   $\mathbf{25.90}$& $\mathbf{55.61}$& $\mathbf{67.66}$ & $\mathbf{89.66}$ & $\underline{55.21}$ & $\underline{75.88}$ & $87.98$& $\mathbf{55.41}$ \\
    \hline \hline
    FashionIQ & Combiner~\cite{baldrati2022effective}& $21.11$& $50.96$& $64.75$& $87.95$& $48.63$& $71.90$& $86.24$&$49.80$\\
    CIRR & Combiner~\cite{baldrati2022effective}& $31.61$&$ 62.22$& $75.23$& $93.52$& $60.63$& $80.84$&$90.99$ & $61.42$\\
    \hline
    \end{tabular}%
\end{table*} 
\vspace{-0.5em}

\subsection{On Model Comparison~(RQ1)}
To comprehensively validate the effectiveness of our method, we adopted the following baselines, including seven zero-shot methods and one classic CLIP-based supervised method. 
\begin{itemize}
    \item \textbf{Image-only.} It simply takes the CLIP features of the reference image to retrieve the target images. 
    \item \textbf{Text-only.} It simply takes the CLIP features of the modification text to retrieve the target images. 
    \item \textbf{Image + Text.} It averages the CLIP features of the reference image and the modification text to retrieve the target image.
    \item \textbf{Captioning.} It first concatenates the caption of the reference image, generated from an image captioning model, and the modification text, and then takes the CLIP features of the composed text to retrieve the target images. For a fair comparison, we adopted BLIP as the image captioning model. 
    \item \textbf{Pic2Word~\cite{saito2023pic2word}.} It employs a textual inversion network to map the reference image into a single pseudo-word token within the CLIP token embedding space. 
    \item \textbf{SEARLE-XL-OTI} and \textbf{SEARLE-XL}~\cite{Baldrati_2023_ICCV}. SEARLE-XL-OTI employs an optimization-based textual inversion to learn a pseudo-word token for encapsulating the visual content of each image, where no mapping function is involved. By distilling knowledge from SEARLE-XL-OTI,  SEARLE-XL learns a compact efficiency mapping network to map the image into a pseudo-word token. Both methods adopt a category-based semantic regularization to align the pseudo-word token to the CLIP token embedding space. 
    \item \textbf{Combiner~\cite{baldrati2022effective}.} This is a standard CLIP-based supervised CIR model, which works on combining the visual and textual features derived from the frozen pre-trained CLIP to fulfill CIR tasks. Due to the absence of the CIRCO training set, we separately trained this model with training splits of FashionIQ (~$18$K triplets) and CIRR (~$28$K triplets). We then evaluated the trained Combiner network on all three datasets, to investigate its generalization ability. 
\end{itemize}

\begin{table}[!t]
  \centering
  \caption{\textbf{Performance comparison on CIRCO.} The best results are in boldface, while the second best results are underlined.}
  \vspace{-0.5em}
  \label{tab: baseline_circo}
  \resizebox{8.5cm}{!}{
  \normalsize
    \begin{tabular}{c|l|cccc}
    \hline
    \multirow{2}{*}{Supervision} & \multirow{2}{*}{Method} & \multicolumn{4}{c}{\textbf{mAP@$K$}} \\
    \cline{3-6}
     & & {$K$ = $5$} & {$K$ = $10$}& {$K$ = $25$}& {$K$ = $50$} \\
    \hline  \hline
    \multirow{8}{*}{ZERO-SHOT} & Image-only & $1.80$& $2.44$& $3.05$& $3.46$\\
    & Text-only& $3.01$& $3.18$& $3.68$& $3.93$\\
    & Image + Text& $4.32$& $5.24$& $6.49$& $7.07$\\
    & Captioning& $8.33$& $8.98$& $10.17$& $10.75$\\
    & Pic2Word~\cite{saito2023pic2word}& $8.72$& $9.51$& $10.46$& $11.29$\\
    & SEARLE-XL-OTI~\cite{Baldrati_2023_ICCV}& $10.18$& $11.03$& $12.72$& $13.67$\\
    & SEARLE-XL~\cite{Baldrati_2023_ICCV}& $\underline{11.68}$& $\underline{12.73}$& $\underline{14.33}$& $\underline{15.12}$\\
    & \textbf{FTI4CIR} &   $\mathbf{15.05}$& $\mathbf{16.32}$& $\mathbf{18.06}$& $\mathbf{19.05}$\\
    \hline \hline
    FashionIQ & Combiner~\cite{baldrati2022effective}& $8.91$& $10.29$& $11.72$&$12.52$\\
    CIRR & Combiner~\cite{baldrati2022effective}& $8.56$& $9.20$& $10.43$& $11.06$ \\
    \hline
    \end{tabular}
    }
      \vspace{-1.0em}
\end{table}

Tables~\ref{tab: baseline_FashionIQ} -~\ref{tab: baseline_circo} summarize the performance comparison on the three datasets. For a fair comparison, we adopted CLIP (ViT-L/$14$ version), as the feature extraction backbone for all baselines. 
From these tables, we obtained the following observations. 
1) On average, FTI4CIR consistently outperforms all zero-shot baselines across three datasets. This reflects the effectiveness of our model in the ZS-CIR setting. 
2) Comparing the results of FTI4CIR with the Combiner network trained on FashionIQ or CIRR demonstrates the generality and effectiveness of FTI4CIR, as well as the domain adaptation challenge in supervised methods. For instance, in Table~\ref{tab: baseline_FashionIQ}, while the Combiner trained on FashionIQ performs better than FTI4CIR, the Combiner trained on CIRR exhibits significantly lower performance. Such a phenomenon is evident across all three provided datasets. This observation highlights domain discrepancies among CIR datasets, and most supervised methods tend to rely heavily on the training triplets, leading to poor generalization on other CIR datasets. 
3) Table~\ref{tab: baseline_cirr} shows that the Text-only baseline achieves the best performance on R$_{subset}$@$K$, which has been previously noted in studies~\cite{baldrati2022effective, saito2023pic2word}. It emphasizes some drawbacks of CIRR: the modification text alone often suffices for image retrieval, rendering the information from reference images redundant. Besides, some reference images can even be harmful to retrieving the target image. These issues are further exacerbated, particularly when the candidate set of the target image exhibits high visual similarity with the reference image, making the semantics of the modification text crucial in such scenarios.

\subsection{On Ablation Study~(RQ2)}
To verify the importance of each component in our model, we compared FTI4CIR with its following derivatives.

    \begin{itemize}
    \item \textbf{w/o_subject_token}. To explore the effect of learning the subject-oriented pseudo-word token, we eliminated the subject-oriented pseudo-word token mapping and accordingly used the template ``a photo with [${[A_1^*,\cdots,A_r^*]}$] but [$T_m$].'' to retrieve the target image.
    \item \textbf{w/o_attribute_token}. To verify the importance of learning the attribute-oriented pseudo-word tokens, we removed these tokens mapping and utilized the template ``a photo of [$S^*$] but [$T_m$].'' to retrieve the target image.
 \item \textbf{w/o_filter}. To show the effectiveness of the local-global relevance-based filtering strategy, we disabled the filter criteria by setting $k=n$ and $\varepsilon = 0$. 
    \item \textbf{w/o_ortho}. To investigate the effect of orthogonal constraint on the local attribute features learning, we removed $\mathcal{L_{\text{orth}}}$.
    \item \textbf{w/o_context_reg}. To verify the benefit of regulating the pseudo-word tokens learning in the full context, we revised the derivative $\hat{T}_S$ to ``a photo of [$S^*$].'', and $\hat{T}_A$ to ``a photo with ${[A_1^*,\cdots, A_r^*]}$.''. 
    \item \textbf{w/o_subject_reg}, \textbf{w/o_attribute_reg}, and \textbf{w/o_whole_reg}. To examine the efficacy of our tri-wise caption-based semantic regularization, we conducted three variants, where the losses $\mathcal{L_{\text{subject}}}$, $\mathcal{L_{\text{attribute}}}$, and $\mathcal{L_{\text{whole}}}$ are removed, respectively.
\end{itemize}

Table~\ref{tab: ablation_study} shows the ablation results of our FTI4CIR on three datasets. From this table, we gained the following observations. 
1) Both \mbox{w/o_subject_token} and \mbox{w/o_attribute_token} perform inferior to FTI4CIR, emphasizing that it is essential to consider both the primary subject(s) and local attributes of the image to handle the diverse modification demands in CIR tasks.
2) FTI4CIR exceeds \mbox{w/o_filter} in various evaluation metrics, which validates the effectiveness of our filtering criteria in getting rid of the irrelevant local attribute features for the given image. 
3) w/o\_ortho performs worse than FTI4CIR, indicating that the designed orthogonal loss indeed helps guarantee the independence among local attribute features and promotes the local attribute features extraction. 
4) w/o\_context\_reg is inferior to FTI4CIR, demonstrating the advantage of conducting the tri-wise caption-based semantic regularization with the whole context rather than the local counterparts. 
5) FTI4CIR achieves better performance than both w/o\_subject\_reg, w/o\_attribute\_reg, and w/o\_whole\_reg, suggesting that each semantic regularization partially contributes to the pseudo-word tokens learning. Besides, compared to other derivatives, w/o\_whole\_reg exhibits the highest performance on FashionIQ, whereas it yields the least favorable outcome on CIRCO. This contrast highlights that the fashion-domain dataset places greater emphasis on the interplay between pseudo-word tokens and real-word tokens, thus effectively conveying user retrieval requirements in comparison to the open-domain dataset. 

\begin{table}[!t]
  \centering
  \caption{Ablation studies on CIRCO, CIRR, and FashionIQ.}
  \label{tab: ablation_study}
  \resizebox{8.5cm}{!}{
    \begin{tabular}{l|cc|c|c}
    \hline
     \multirow{2}{*}{Method} & \multicolumn{2}{c|}{FashionIQ-Avg} & \multicolumn{1}{c|}{CIRR} & \multicolumn{1}{c}{CIRCO} \\
    \cline{2-5} 
     & {\textbf{R@$10$}} & {\textbf{R@$50$}} & Avg & {\textbf{mAP@$5$}}\\
     \hline \hline 
    {w/o_subject_token}& $19.83$& $37.11$& $54.18$& $8.36$\\
    {w/o_attribute_token}& $27.72$& $48.79$ & $51.22$&$12.69$\\
    {w/o_filter}& $23.13$& $42.63$ & $46.07$&$11.34$\\
    {w/o_ortho}& $25.51$& $45.33$ & $54.13$& $14.57$\\
    {w/o_context_reg}& $24.96$& $44.16$& $48.81$& $10.87$\\
    {w/o_subject_reg}& $25.74$& $45.61$ &$54.54$ &$14.18$\\
    {w/o_attribute_reg}& $25.15$& $46.30$ &$52.28$ &$12.61$\\
    {w/o_whole_reg}& $26.39$& $46.87$ &$53.17$ & $11.65$\\
    \textbf{FTI4CIR}& $\mathbf{29.39}$& $\mathbf{50.88}$ & $\mathbf{55.41}$&$\mathbf{15.05}$\\
    \hline
    \end{tabular}
    \vspace{-1.0em}
    }
\end{table}

\begin{figure}[!t]
		\centering
		\includegraphics[scale=0.41]{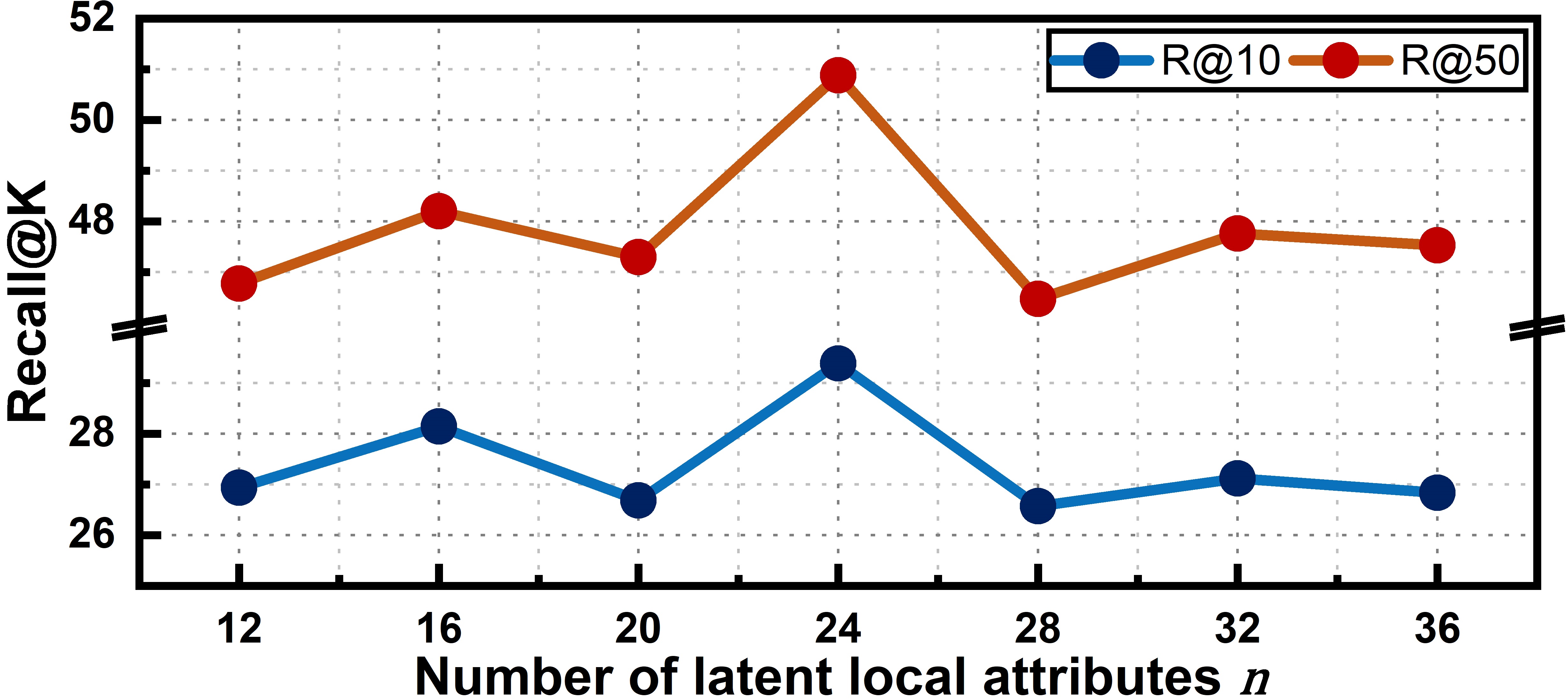}
   \vspace{-1.5em}
	    \caption{Sensitivity analysis of our model on the number of latent local attributes $n$. Notably, we reported the average results of R@$10$ and R@$50$ on FashionIQ.}\label{fig:sensitivity_figure}
       \vspace{-1.5em}
\end{figure}

\subsection{On Sensitivity Analysis~(RQ3)} 
In this part, we tested the sensitivity of our model regarding the number of latent local attributes, \textit{i.e.}, $n$. As shown in Figure~\ref{fig:sensitivity_figure}, we varied the number of latent local attributes $n$ from $12$ to $36$ at the step of $4$. As can be seen, the performance of our model generally boosts with the increasing number of latent local attributes, followed by a subsequent decline. This is reasonable, as a greater number of latent local attributes can differentiate the latent attribute features of an image in more detail and hence facilitate the interaction between the reference image and the modification text; while an excessive number of latent local attributes may lead to excessive differentiation of local attribute features and make it challenging for the filter to screen out the irrelevant attribute features.

\subsection{On Case Study~(RQ4)}
\begin{figure}[!t]
		\centering
		\includegraphics[scale=0.62]{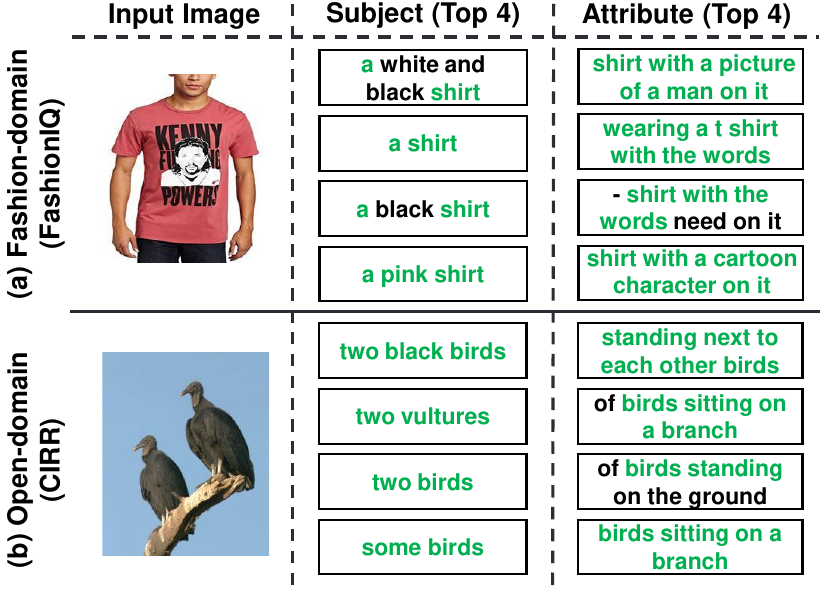}
  \vspace{-1.5em}
	    \caption{Pseudo-to-real description retrieved results. We highlight the related real-word descriptions in green.} \label{fig:case_study1}
  \vspace{-1.5em}
\end{figure}
In this part, we show illustrative results on tasks of pseudo-to-real description retrieval and composed image retrieval.

\subsubsection{Pseudo-to-Real Description Retrieval.} 
To showcase that the learned subject-oriented and attribute-oriented pseudo-word tokens can accurately express the primary subject(s) and local attributes of the image, respectively, we conducted the pseudo-to-real description retrieval. In particular, we first regarded all the BLIP-generated $\hat{T}_{subj}$/$\hat{T}_{attr}$ for images in the pre-training dataset as the candidate set of real-word-based subject/attribute descriptions. We then used the image's pseudo-word-based textual (``a photo of [$S^*$] with [${[A_1^*,\cdots,A_r^*]}$].'') representation to retrieve the related subject and attribute descriptions from the above candidate set, respectively. 
Figure~\ref{fig:case_study1} shows the top-$4$ retrieved descriptions for two testing cases, where the related real-word descriptions are highlighted in green. As can be seen from Figure~\ref{fig:case_study1}~(a), the retrieved subject descriptions are correctly concerned about ``a shirt'', while the attribute descriptions describe the pattern of the shirt. Regarding Figure~\ref{fig:case_study1}~(b), the retrieved subject descriptions correctly reflect the number and type of the primary subjects, namely ``two birds'', and the attribute descriptions reveal the position and spatial information of the two birds. Overall, these results suggest the effectiveness of the learned pseudo-word tokens in capturing the image content and the alignment between the learned pseudo-word tokens to the real-word token embedding space.

\subsubsection{Composed Image Retrieval.} 
\begin{figure}[!t]
		\centering
		\includegraphics[scale=0.41]{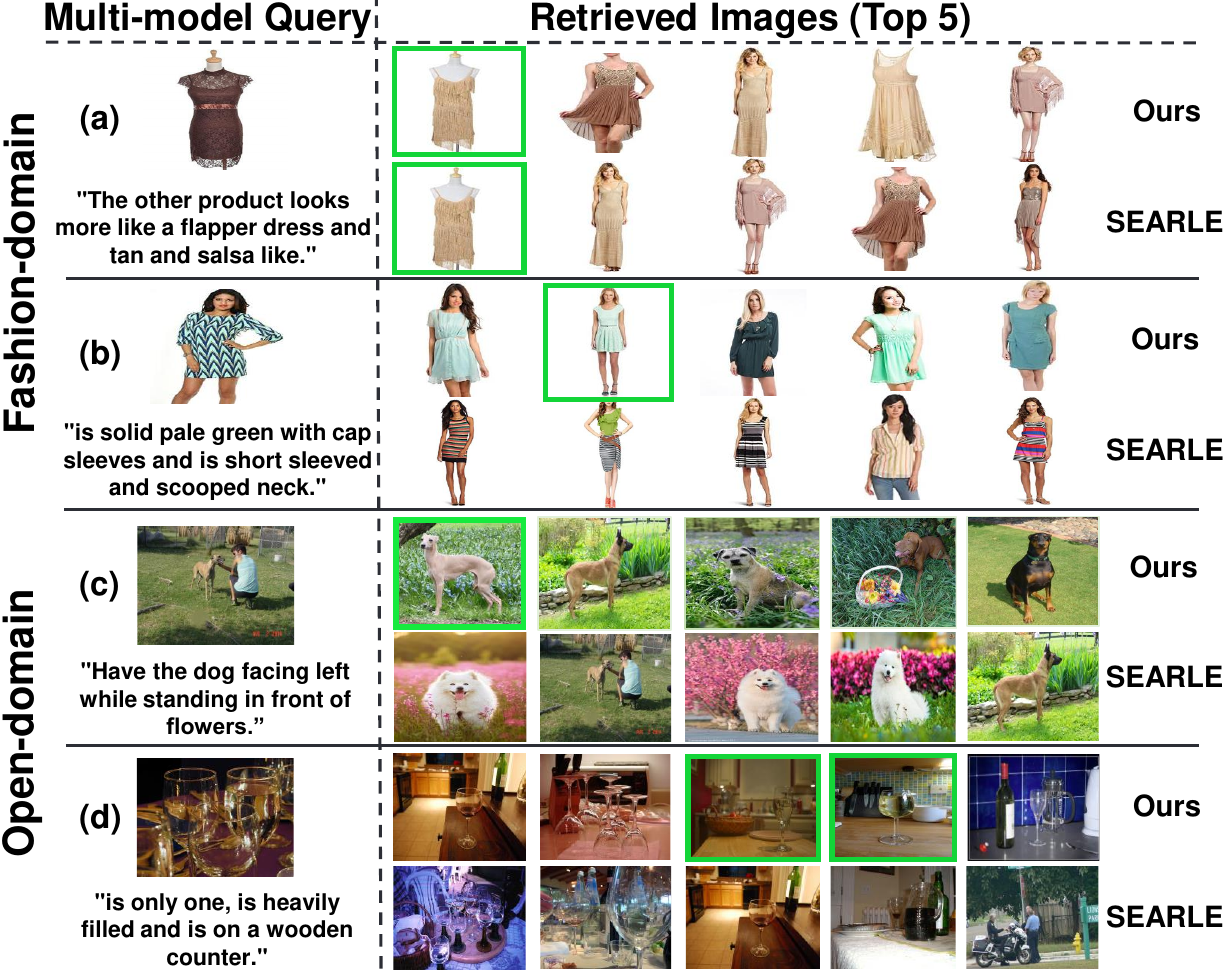}
   \vspace{-1.5em}
	    \caption{Illustration of CIR results on three datasets. (a) and (b) come from FashionIQ, (c) comes from CIRR, and (d) comes from CIRCO. The ground-truth target images are highlighted in green boxes. }\label{fig:case_study2}
   \vspace{-2.0em}
\end{figure}

Figure~\ref{fig:case_study2} illustrates several CIR results obtained by our FTI4CIR and the best-performing baseline SEARLE-XL on the fashion-domain dataset FashionIQ and the open-domain dataset CIRR and CIRCO. Due to the limited space, we reported the top $5$ retrieved images. As can be seen from the first case, shown in Figure~\ref{fig:case_study2}~(a), the user wants to change the global visual properties of the given dress. For this modification request, both our FTI4CIR and SEARLE-XL correctly rank the target image in the first place. Nevertheless, for more complex cases in Figure~\ref{fig:case_study2}~(b) and~(c), the modification texts primarily center on local attribute alterations to the reference images. The former involves adjusting \textit{sleeve length} and \textit{collar type}, while the latter emphasizes modifying the \textit{posture} and \textit{background}. In such cases, our FTI4CIR succeeds in ranking the ground-truth target images in the top-$2$ places, while SEARLE-XL fails to rank them within the top-$5$ places. This implies the superiority of the fine-grained textual inversion over the conventional coarse-grained textual inversion. Last but not least, as illustrated in Figure~\ref{fig:case_study2}~(d), the primary subjects of the reference image are multiple cups, while the modification request involves not only detailed attribute changes, such as ``heavily filled'' and ``on a wooden counter'', but also the change of the number of subjects, \textit{i.e.}, ``only one''. For this case, our model still outperforms SEARLE-XL. This may be attributed to the tri-wise caption-based semantic regularization, in which both the subject-oriented and attribute-oriented pseudo-word tokens are enhanced by specific regularization. Notably, in our work, the subject-oriented pseudo-word is designed to capture both the category and the number of the primary subject(s) in the image.

%% file: sec_concl.tex
\section{Conclusions and Future Work}
\label{sec:concl}
In this work, we propose a novel fine-grained textual inversion network to tackle the challenging task of ZS-CIR. Beyond current solutions, we map each image into a subject-oriented \mbox{pseudo-word} token and several attribute-oriented \mbox{pseudo-word} tokens to encapsulate the image content in a textual sentence effectively. In addition, we design a tri-wise caption-based semantic regularization to boost the alignment of the fine-grained pseudo-word tokens with the real-word token embedding space. 
Extensive experiments have been conducted on three public datasets, and the results demonstrate the effectiveness of our method. 
In addition, ablation studies verify the effectiveness of each key component of our model, confirming the benefit of conducting fine-grained textual inversion and tri-wise caption-based semantic regularization. 
To take it further, we  plan to extend our method to address the challenging multi-turn interactive image retrieval task in a zero-shot manner. We believe that our fine-grained textual inversion network could fulfill the diverse retrieval requirements of different users.